\documentclass[10pt, twocolumn]{IEEEtran}

\ifCLASSINFOpdf
  \usepackage[pdftex]{graphicx}
\else
\fi

\usepackage{amsfonts,amsmath,amsthm,amssymb,graphicx,algorithm,epstopdf,cite,url,subfigure}
\usepackage{psfrag,amssymb,amsmath,pifont,cite,graphics,graphicx,epsfig,subfigure,amscd,helvet,multirow,enumerate}
\usepackage{lmodern}
\usepackage{color}
\usepackage{epstopdf}
\usepackage{accents}
\usepackage{amssymb,amsmath,amscd,cite,enumerate}
\usepackage{graphics,graphicx,epsfig,psfrag,subfigure,threeparttable,url}
\usepackage[colorlinks,citecolor=black,linkcolor=black]{hyperref}
\usepackage{algpseudocode}
\usepackage[english]{babel}
\usepackage{graphicx}
\usepackage{amsmath,amssymb} 
\usepackage{color}
\usepackage{epsfig}
\usepackage[table]{xcolor}
\usepackage{adjustbox, enumerate}
\usepackage{multirow}
\usepackage{makecell}
\usepackage{caption}
\usepackage{wrapfig}
\usepackage{lipsum}
\usepackage{subfigure}
\usepackage{cite}
\usepackage[misc]{ifsym}

\graphicspath{{figures/}}

\begin{document}

\captionsetup[figure]{labelformat={default},labelsep=period,name={Fig.}}

%
\title{ADCC: An Effective and Intelligent Attention Dense Color Constancy System for Studying Images in Smart Cities}


\author{
Yilang Zhang, Neal N. Xiong, Zheng Wei,  Xin Yuan and Jian Wang
\thanks{Y. Zhang, Z. Wei and J. Wang are with the School of Data Science, Fudan University, Shanghai 200433, China (e-mail: \{yilangzhang16, zwei19,  jian\_wang\}@fudan.edu.cn).}
\thanks{N. Xiong is with Department of Mathematics and Computer Science,
Northeastern State University, Tahlequah, OK, USA (e-mail: xiongnaixue@gmail.com, xiong31@nsuok.edu).} 
\thanks{X. Yuan is with Nokia-Bell Labs, New Providence, NJ 07974-0636 USA (e-mail: xyuan@bell-labs.com).}
}

%




\IEEEtitleabstractindextext{%
\begin{abstract}
As a novel method eliminating chromatic aberration on objects, computational color constancy has becoming a fundamental prerequisite for many computer vision applications. Among algorithms performing this task, the learning-based ones have achieved great success in recent years. However, they fail to fully consider the spatial information of images, leaving plenty of room for improvement of the accuracy of illuminant estimation. In this paper, by exploiting the spatial information of images, we propose a color constancy algorithm called Attention Dense Color Constancy (ADCC) using convolutional neural network (CNN). Specifically, based on the $2$D $\log$-chrominance histograms of the input images as well as their specially augmented ones, ADCC estimates the illuminant with a self-attention DenseNet. The augmented images help to tell apart the edge gradients, edge pixels and non-edge ones in $\log$-histogram, which contribute significantly to the feature extraction and color-ambiguity elimination, thereby advancing the accuracy of illuminant estimation. Simulations and experiments on benchmark datasets demonstrate that the proposed algorithm is effective for illuminant estimation compared to the state-of-the-art methods. {Thus, ADCC offers great potential in promoting applications of smart cities, such as smart camera, where color is an important factor for distinguishing objects.}
\end{abstract}

}
\maketitle
 

\IEEEdisplaynontitleabstractindextext

%
\IEEEpeerreviewmaketitle

\section{Introduction}
{\it This work has been submitted to the IEEE for possible publication. Copyright may be transferred without notice, after which this version may no longer be accessible.}
\IEEEPARstart{C}{o}lor is an essential cue for studying images. The color reflected in an image is determined by the intrinsic properties of objects, surfaces and light sources~\cite{summary}. {To obtain the color of objects under the standard light source (i.e., the white light), one has to eliminate the chromatic aberration caused by the light sources,  which constitutes the goal of computational color constancy.} 
Over the years, computational color constancy has been a long-standing problem in many fields, such as visual science and computer vision. While many existing methods have been demonstrated to be effective in solving this problem, there are still challenges in both accuracy as well as computational efficiency.

\begin{figure}[t]
	\centering
	 \includegraphics[width=1.0\linewidth]{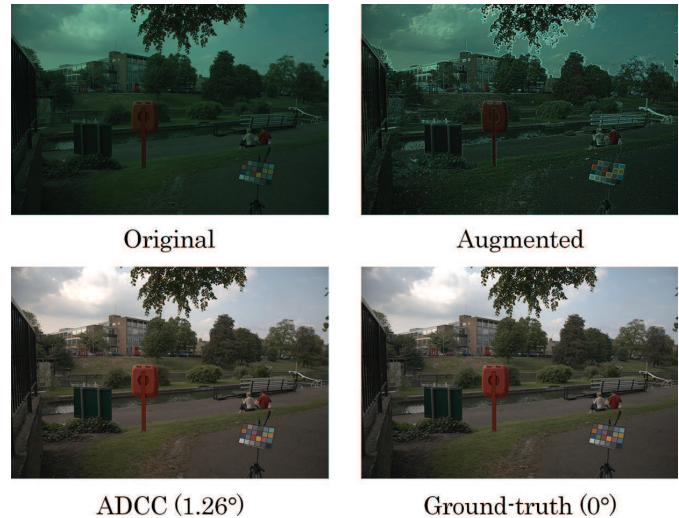}
    \caption{An illustrative example of an image restored by the ADCC algorithm. Top-left: the original image; top-right: the augmented image of the original one; bottom-left: the image restored by ADCC (with angular error $1.26^\circ$); bottom-right: the image restored by the ground-truth illuminant.}
	\label{fig:intro}
\end{figure}

Existing color constancy methods generally assume some regularities for the color of a natural object observed under the white light. For example, the most simply designed Gray World~\cite{GW} algorithm uses the assumption that the average reflectance in a scene is achromatic under a natural light source. Also, the arithmetic mean was generalized via higher order mathematical calculations~\cite{edgebased}. These hypothesis-based algorithms are often classified as learning-free algorithms. Recently, with the innovative deep learning networks making a splash in computer vision, they have greatly promoted the research of learning-based methods for color constancy. Indeed, since the paper~\cite{CNN} debuted in 2015, which firstly used the convolutional neural network (CNN) to solve the color constancy problem, there has been a growing number of works that used deep learning frameworks for this task~\cite{FC4, CCC, FFCC}. Notably, many of them have achieved satisfactory performance on the standard color constancy benchmark datasets~\cite{Shi, Cheng}.  


In this paper, we present a color constancy algorithm dubbed Attention Dense Color Constancy (ADCC). It is inspired by the recent successful approaches~\cite{CCC, FFCC} that reformulate the color constancy problem as a two-dimensional ($2$D) spatial localization task in the $\log$-chrominance space. 
\begin{itemize}
\item While such formulation largely simplifies the problem by reducing the number of parameters to be estimated, it also raises the issue that the {\em spatial information} of input images, which usually offers significant cues for illuminant estimation, is somehow {\em missing} in the $\log$-chrominance space. 
\item It becomes difficult, when transforming into the $\log$-chrominance space, to distinguish ambiguous pixels in an image. The ambiguous pixels, e.g., the {\em edges} or shadows of an image, are not directly related to the illuminant, but could severely interfere with the illuminant estimation.
\end{itemize}

To address these issues, we propose an edge augmentation operator, which can well preserve the spatial information of images and also eliminate the color ambiguity of edges, when translating into the $\log$-chrominance space. Specifically, it makes better use of the spatial information through extracting the gradients of edges, while retaining the non-edge pixels of an original image in its augmented counterpart. We thus feed the original images together with their augmented ones (both in $\log$-histogram) as inputs into a {\em self-attention DenseNet} for illuminant estimation. An image restored by ADCC is shown in Fig.~\ref{fig:intro}.


In fact, not only can our network take full advantage of the spatial information of the original image, it also confers many other merits, such as the end-to-end training, adaptive processing for images with arbitrary sizes, and robustness of the final estimation. Through experiments on the reprocessed Color Checker dataset~\cite{Shi2000re}, it is demonstrated that the ADCC algorithm achieves competitive performance compared to the state-of-the-art methods. Meanwhile, it is also flexible enough to reach satisfactory performance on the NUS $8$-camera dataset~\cite{Cheng}.    

 The rest of this paper is organized as follows: Section~II introduces the related work on the computational color constancy. Section III introduces the proposed ADCC algorithm. Section IV presents the simulation and experimental results that demonstrate the effectiveness of ADCC. Section V concludes out work and discusses some future directions.

\section{Related Work}
 
Over the years, {there have much lots of researches on the color of objects~\cite{xiong2018color, hu2008color,yang1999color}. As one of the most important topics in this field, computational color constancy has received much attention.}  Existing methods can be roughly grouped  into two major categories: i) the learning-free and ii) learning-based frameworks. Typically, the former assumes some particular statistical  or  physical priors of natural images. 
In~\cite{GW}, Buchsbaum proposed the well-known gray-world assumption that any deviation from achromaticity in the average scene color is caused by the effects of the illuminant, which means that the color of the light source can be estimated by computing the average color in the image. Land put forward another famed assumption in~\cite{wp}: {\em the maximum response in the RGB-channels is caused by a perfect reflectance.} In other words,  a surface with perfect reflectance properties will reflect the full range of light that it captures. Therefore, the illuminant estimation can be obtained by computing the maximum response in the separate color channels. In~\cite{sog}, the white patch and gray-world algorithms were shown to be special instantiations of the more general Minkowski framework. Weijer {\it et al.}~\cite{edgebased} proposed the incorporation of higher order image statistics in form of derivatives. Moreover, this method has been further enhanced and improved in~\cite{chen2007edge, chakrabarti2008color, gijsenij2009physics}.

In addition to Minkowski framework, Forsyth~\cite{forsyth1990novel} introduced another type of learning-free method. It is based on the assumption that in real-world images, for a given light source, only a limited number of colors can be observed. Consequently, any variations in the colors of an image are caused by the color deviation of the light source. To estimate the illuminant in the colored picture, this method employs a restraint system. These constraints stem from physical restrictions in form of surface reflectance functions. The variants and extensions of this method can be found  in~\cite{finlayson1996color, finlayson2000improving, finlayson1999selection}.  

Methods in this category are irrelevant to the dataset and camera information and also have no need of training. 
However, due to the imprecision of priors and random fluctuation of statistics in varieties of practical images, which often lead to the biased and noisy results, the learning-free methods might not be able to achieve satisfactory performance. 

In the second category, methods aim at constructing a model from the training data, which essentially search over the entire assumption space for the best prior. Early learning-based methods often employed some simple structures and algorithms for illuminant estimation. In~\cite{SVM,simplefeature,simpleuni}, for instance, some plain features were extracted manually to regress the illuminant prediction with linear regression or support vector machine (SVM).  In~\cite{exemplar},  nearest neighbor methods were employed to solve the illuminant prediction problem. While these methods have demonstrated great advantages over the learning-free ones, their manufactured features still fail to characterize the whole information of images. 

To better extract the information of images, Bianco {\it et al.}~\cite{CNN} first utilized the CNN for semantic feature extraction in illuminant estimation.  {CNNs have been widely employed in image detection, recognition and other computer vision tasks such as~\cite{muhammad2018efficient, kamel2018deep, wang2018re, tao2018detection}.} Since then, many variants of CNNs have been employed to perform this task. For example, Shi \textit{et al.} \cite{DSNet} proposed a two-branch structure to generate prediction from two illuminant hypotheses. \cite{objrecog} determined the illuminant from objects whose colors are learnt through object recognition. \cite{FC4,quasi} took advantage of image segmentation, which assigns different weights to a mask for different pixels of the image. In~\cite{CNN,singmul,DSNet}, the patch-based methods were also utilized to obtain the global estimation from local candidates. In summary, by exploiting the spatial structure of images that contains significant semantic information, the CNN-based methods have demonstrated significant improvement over the previous ones. However, it should also be noted that these methods essentially perform under the multiplicative constraints~\cite{CCC}, which might be too complex for some practical scenarios. 

There have been some recent efforts in addressing the multiplicative constraints. For example, Barron~\cite{CCC,FFCC} reformulated the color constancy problem in a much simpler way. To be specific, images represented in RGB channels are transformed to the $2$D $\log$-chrominance histograms. Interestingly, this operation reduces the number of parameters to be estimated from three to two, thereby greatly simplifying the problem. Moreover, since the varying of illuminant is  equivalent to the linear shift in the space of $\log$-chrominance, the multiplicative constraints are naturally translated into linear ones~\cite{CCC}.


\section{The Proposed ADCC Algorithm}

\subsection{Overview}
\paragraph{Problem formulation}
For an RGB image, the image value of pixel $k$ under the Lambertian assumption follows:
\begin{align}
I_c ^{k} = \int_\omega L^{k}(\lambda) \rho(\lambda) S^{k}(\lambda) d \lambda,\quad c\in \{r, g ,b\},
\label{formation}
\end{align}
where $\lambda$ is the wavelength of the light that belongs to the range $\omega$, $L^{k}(\lambda)$ is the light source to be estimated at pixel $k$, $\rho(\lambda)$ is the sensitivity function of the camera, and $S^{k}(\lambda)$ is the surface reflectance of the specific object at pixel $k$. 

For a single light source, it is assumed that the illuminant has constant values: 
\begin{equation}
 L^{k}(\lambda) \equiv(L_r,L_g,L_b),
 \end{equation} on three channels for all pixels of the image. Then, Eq.~\eqref{formation} reduces to a diagonal model called the~\textit{von Kries Model}~\cite{vonKries}. The goal of color constancy becomes to correct the image to what it should be under the canonical illuminant:
\begin{equation}
 L_{\text{canon}}= \left(\frac{1}{\sqrt{3}},\frac{1}{\sqrt{3}},\frac{1}{\sqrt{3}} \right),
 \end{equation} 
where $L_{\text{canon}}$ is normalized because the correction is only for the \textit{chrominance}, rather than for the brightness level~\cite{summary}. In a nutshell, the color constancy methods perform two steps: i) estimating the illuminant for RGB channels from an image and ii) correcting each pixel of the image by eliminating the effect caused by the estimated illuminant.

\paragraph{Metric}
Following previous works (e.g.,~\cite{CNN,CCC}), we use the three-fold cross validation to evaluate the performance of our method. 
	 For an input image, the evaluation criteria are based on the angular error (in degree) introduced by Hordley and Finlayson~\cite{metric}. In particular, the angle between the RGB triplet of the estimated illuminant $\hat L$ and that of the measured ground-truth illuminant $L$ is given by
\begin{align}
\label{metric}
E(L, \hat L) = \frac{180^\circ}{\pi} \arccos \left(\frac{L^T\hat L}{\Vert L\Vert_2 \Vert \hat L\Vert_2}  \right).
\end{align}
We consider the following important metrics: i) mean, ii) median, iii) tri-mean of all the errors, iv) mean of the lowest $25\%$ errors,  v) mean of the highest $25\%$ errors, and vi) $95\%$ quantile error, which have been commonly used in the color constancy literatures.

\paragraph{Transformation to $\log$-chrominance}
Following the popular transformaton in~\cite{CCC}, we project the images onto the UV space.  The $\log$-chrominance \textit{u} and \textit{v} for pixel $k$ of the original image are defined as:
\begin{align} \label{eq:Iuv}
	I_u ^{k} = \log \left ( \frac{I_r ^{k} }{I_g ^{k}} \right) 
\end{align}
and
\begin{equation}
I_v ^{k} = \log \left ( \frac{I_b ^{k}}{I_g ^{k} } \right),
\end{equation}
respectively. Likewise, the illuminant in the UV space, whose scale we are not concerned about, is given by:
\begin{align}
	L_u  = \log \left ( \frac{L_r}{L_g} \right)
\end{align}
and 
\begin{equation}
L_v = \log \left ( \frac{L_b}{L_g} \right),
\end{equation}
respectively. With the estimation of $L_u$ and $L_v$, the normalized RGB illuminant can be recovered through:
\begin{equation} \label{recover_illum}
L_r = \frac{\exp{\left(L_u\right)}}{z},~~ L_g=\frac{1}{z}~~\text{and}~~ L_b=\frac{\exp{\left(L_v\right)}}{z},
\end{equation}
where $$z=\sqrt{\exp{ (L_u )}^2+\exp{ (L_v )}^2 + 1}.$$

From Eq.~\eqref{eq:Iuv}, the chrominance of all pixels in an image can be put into a histogram comprised of small bins. In this histogram, $M_I(u,v)$ counts the number of pixels in image $I$ whose chrominance is close to the values $(u,v)$:
\begin{equation} \label{eq:MI}
M_I(u, v)=\sum_{k \in I}\left[\left\vert I_{u}^{k}-u\right\vert \leq\frac{\epsilon}{2}\wedge\left\vert I_{v}^{k}-v\right\vert \leq\frac{\epsilon}{2}\right], 
\end{equation}
where $\epsilon$ controls the size of each bin in the histogram. Thus, the multiplicative change of illuminant in the RGB space can be translated to the additive change of illuminant in the UV space. To estimation the illuminant of an input image, we only need to detect the linear shift of $\log$-chrominance in the $2$D $\log$-histogram.

\begin{figure*}[!t]
		\centering
			\includegraphics[width=165mm]{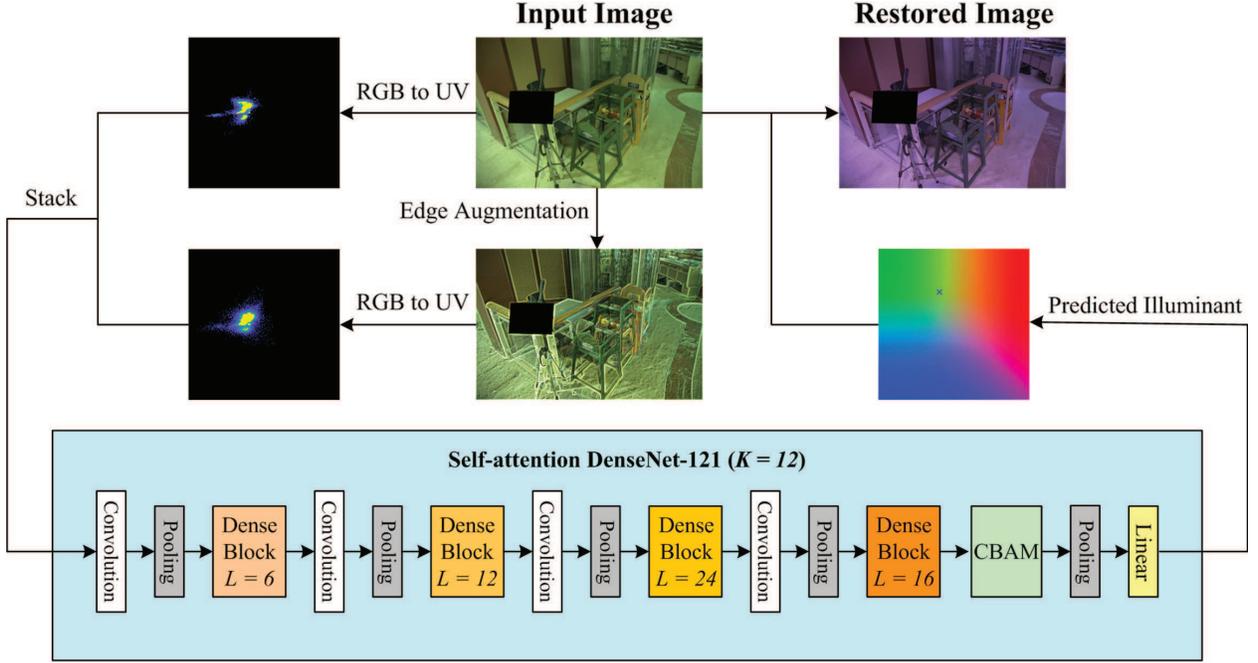} 
		\caption{The network architecture of DCC.}
		\label{fig:architecture}
	\end{figure*}

\paragraph{Method}
The major steps of the proposed algorithm are summarized as follows. We first perform the edge augmentation for each image. Next, by translating the images and their augmented edges to the UV space, we stack them as two channels of inputs to our network. Then, we employ a self-attention DenseNet to estimate the illuminant for each input, where the self-attention is employed to reweight the feature maps. Finally,  from the network output $(L_u,L_v)$, we obtain the illuminant estimation; see Eq.~\eqref{recover_illum} for details. The chart of our ADCC algorithm is shown in Fig.~\ref{fig:architecture}.


\subsection{Edge Augmentation and Utilization}
\label{sec:edge}
For years, edge augmentation methods have been extensively used for data augmentation. For those performing in the RGB space~\cite{FC4, quasi}, edges are often augmented to help segment images. Whereas for those performing in the UV space~\cite{CCC,FFCC}, the augmented edges can be used to measure the local gradients of images, which reflect part of spatial statistics. For example,~\cite{FFCC}  utilized the augmented edges to the network and improved the performance.  

In contrast to the previous researches that abandon non-edge pixels (internal color blocks) when extracting edges, we augment the edges while preserving the non-edge pixels. Later on, we shall give in Table~\ref{table1} a circumstantial evidence for the advantage of retaining non-edge pixels, where the ADCC algorithm indeed performs worse when employing FFCC's approach to edges~\cite{FFCC} for instead. We would like to mention that some recent works also took advantage of the non-edge pixels for illuminant estimation, see, e.g.,~\cite{gao, akbarinia}. They manually assigned weights to adjust the contributions of edge and non-edge pixels. Whereas, we just let the network learn the respective contributions.



We now proceed to introduce our approach to edges. In image processing and computer vision~\cite{sobel_det_1, sobel_det_2}, Sobel filter \cite{sobel} has been widely used for edge extraction for its accurate estimation of gradients and inexpensive computation. To preserve the non-edge pixels, we use the modified Sobel filters as our edge augmentation operator:
\begin{align}
\label{eq:operator}
\begin{split}
	f_x&=
	\begin{bmatrix}
		-1 & 0 & 1\\
		-2 & \sigma & 2\\
        -1 & 0 & 1
	\end{bmatrix},
\end{split}
\end{align}
and 
\begin{equation}
f_y=f_x^T,
\end{equation}
which extract the gradient information of images in horizontal $x$ and vertical $y$, respectively. The intensity and angle of  the augmented edge for $c \in \{r,g,b\}$ are
\begin{equation} \label{eq:ec}
E_c = \sqrt{(f_x\ast I_c)^2 + (f_y \ast I_c)^2} 
\end{equation}
and
\begin{equation}
\Theta_c = \arctan \left( \frac{f_y\ast I_c}{f_x\ast I_c} \right),
\end{equation}
respectively, where ``$\ast$'' denotes the convolution operator. To ensure that our augmented edges are robust to image rotation, only the intensity part is retained as the final augmented edge.
We stress that the hyper-parameter $\sigma$ in Eq.~\eqref{eq:operator} controls the relative proportion between the gradient information and internal color block information of images to be extracted. 
It is easily  seen that the modified operator reduces to the conventional Sobel filter~\cite{sobel} when $$\sigma =0.$$

Furthermore, we can split the operator into two parts:
\begin{equation}
f_x = f_x^{\text{grad}} + f_x^{\text{img}}, 
\end{equation}
where
\begin{equation}
f_x^{\text{grad}} = 
	\begin{bmatrix}
		-1 & 0 & 1\\
		-2 & 0 & 2\\
        -1 & 0 & 1
	\end{bmatrix}
\end{equation}
and
\begin{equation}	
	 f_x^{\text{img}} = 
		\begin{bmatrix}
		0 & 0 & 0\\
		0 & \sigma & 0\\
        0 & 0 & 0
	\end{bmatrix} 
\end{equation}
represent the gradient-component and image-component operator, respectively. 
When the modified operator is applied to the pixels located at the edge in the vertical direction, it is clear that $
 f_x^{\text{grad}} \ast I_c$ dominates  $f_x^{\text{img}} \ast I_c$. Hence,
\begin{equation}
  f_x \ast I_c \approx f_x^{\text{grad}} \ast I_c,
  \end{equation} 
  which actually becomes the local gradient estimation. 
Similarly, when this operator works on the non-edge pixels, we have 
\begin{equation} \label{eq:signaI}
 f_x \ast I_c \approx f_x^\text{img} \ast I_c= \sigma I_c,
 \end{equation} 
 i.e., the preserved non-edge pixels scaled with a scalar. Since the color of edge pixels is the mixture and blur of their surroundings, it can be viewed as the noise. Whereas, the gradient of edge pixels provides us with the spatial and distributional information. The same rationale also applies to the operator $f_y$.

To utilize the edge information and at the same time get rid of the noise brought in by the edge, we transform the image and its augmented edge into two $\log$-histograms:
\begin{equation}
 \hspace{-.5mm}  \hspace{-.5mm}  \hspace{-.5mm} M_{\text{img}}(u, v)  =  \hspace{-.5mm} \sum_{k \in \text{img}}   \hspace{-.5mm} \left[\left\vert I_{u}^{k}-u\right\vert \leq\frac{\epsilon}{2}\wedge\left\vert I_{v}^{k}-v\right\vert \leq\frac{\epsilon}{2}\right],~\hspace{-2mm}
\end{equation}
and 
\begin{equation}
 \hspace{-1mm} M_{\text{edge}}(u, v)  \hspace{0mm} =  \hspace{-.5mm} \sum_{k \in \text{edge}}   \hspace{-.5mm} \left[\left\vert I_{u}^{k}-u\right\vert \leq\frac{\epsilon}{2}\wedge  \hspace{-.5mm} \left \vert I_{v}^{k}-v\right \vert  \hspace{-.5mm}  \leq \frac{\epsilon}{2}\right],~\hspace{-2mm}
\end{equation}
 which act as two channels of the network inputs. Different from~\cite{CCC,FFCC}, whose channels are parallel and irrelevant, our channels are interrelated owing to the dense-connection structure of our network. Ideally, when the chrominance of edge pixels and the non-edge pixels are not overlapped in the $\log$-histogram, the following features can be readily taken advantage of:

\begin{itemize}
\item $\min \left \{M_{\text{img}}(u,v),  M_{\text{edge}}(u,v) /{(\sqrt {2} \sigma}) \right \}$: The common chrominance of the image and its edge corresponds to the non-edge pixels in the RGB space, which we preserve in our augmented edge channel. The denominator $\sqrt {2} \sigma$ is a normalization factor derived from Eqs.~\eqref{eq:ec} and~\eqref{eq:signaI}.

\item $\max \left \{M_{\text{edge}}(u,v) /{(\sqrt {2} \sigma)} - M_{\text{img}}(u,v) ,0 \right \}$: After removing  the chrominance of non-edge pixels from $M_{\text{edge}}$, we obtain the chrominance of the augmented edge pixels, which reflects the statistical information of gradients in edges.

\item $\max \left \{M_{\text{img}}(u,v) -  M_{\text{edge}}(u,v) /{(\sqrt {2} \sigma)}, 0 \right \}$: In a similar way, after removing the chrominance of non-edge pixels from $M_{\text{img}}$, what remains is the chrominance of edges that is not augmented, i.e., that of the noisy pixels to be eliminated.

\end{itemize}

When using hand-crafted features, we may ignore some sophisticated or effective ones. Instead, we just  put $M_{\text{img}}$ and $M_{\text{edge}}$ into the network and  let it learn the features by itself. Since the two chrominances are more or less overlapped, they may lead to some untrustworthy features. To ease the effect of these untrustworthy features, we introduce a self-attention module to our network, which will be discussed in Section~\ref{sec:selfatt}. 

\subsection{The Network Architecture}
A sketch of the ADCC network is provided in Fig.~\ref{fig:architecture}. The network follows a similar structure as in DenseNet-$121$~\cite{dnn}, which has been widely used for solving computer vision problems. In particular, we draw inspiration from this network due to the following two reasons. First of all, according to the design of DenseNet-$121$, the feature-maps generated from the preceding layers can be passed through all the subsequent layers. Even if some of the features are lost during the propagation process, they can still be regenerated at the input of the latter layers through the dense connections~\cite{regenerated}. Note that  the shallow features play an important role in illuminant estimation. Thus, DenseNet-$121$ can be well suited for extracting the features of images in this task.

Secondly, the feature maps from different layers have various receptive fields. Since each layer can get feature maps from all preceding layers through dense connections, the final feature outputs become more effective after aggregating different sizes of receptive fields. Moreover, since DenseNet-$121$ implements shorter connections, and also because we set the growth rate $K= 12$, the feature maps of each layer are relatively small so that DenseNet-$121$ can complete this task with fewer parameters.

DenseNet-$121$ has four dense blocks, for which the numbers of the dense layers are $6$, $12$, $24$ and $16$, respectively. For a dense layer, it contains the same sequence operations: Batch Normalization, ReLU function, and Convolution. The last layer of the network is a fully connected layer. 
We do not directly use the evaluation metric in Eq.~\eqref{metric} as the loss function, since otherwise
\begin{equation}
\lim_{\hat L \to L} \frac{\partial}{\partial \hat L}\left(\arccos  \frac{L^T\hat L}{\Vert L\Vert_2  \Vert \hat L\Vert _2}  \right) = \infty,
\end{equation}
i.e., the gradient value may overflow when the prediction approaches the ground-truth.
Instead, we define the loss function $\mathcal{L}$ of the network in terms of the cosine of the angle between the network estimation $\hat L$ and the ground-truth $L$:
\begin{equation}
    \mathcal{L} :=1 -\frac{L^T\hat L}{\Vert L\Vert_2 \Vert \hat L\Vert_2 },
    \end{equation}
whose derivative is finite and thus does not overflow.

\begin{figure}[t] \centering  
\includegraphics[width=0.45\textwidth]{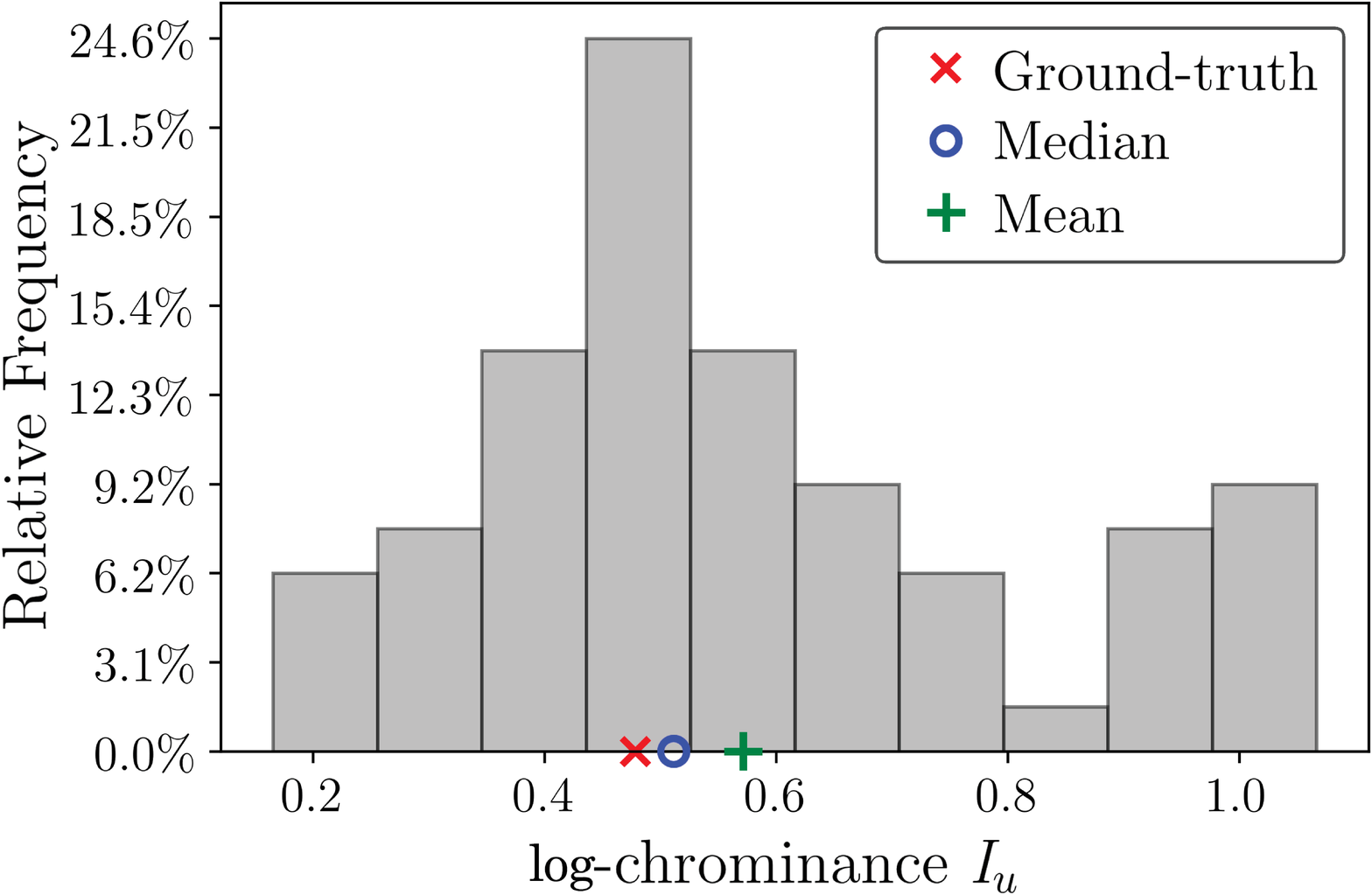}
\caption{An illustrative example for the mean and median of local candidates and the ground-truth illuminant, where ADCC  is used to predict the illuminant $\left( I_u, I_v \right)$ in the UV space for the input image and its $64$ random patches. Due to the outliers of $I_u$ shown in the right-hand side of the histogram, the mean of local candidates deviates more severely from the ground-truth than the median one. We thus use the median of local candidates as our global prediction.} \label{fig:median}
\end{figure}

\subsection{Self-attention} \label{sec:selfatt}
Attention helps to increase the representation power of images by highlighting critical features while suppressing unnecessary ones. In general, the attention module for images considers both the spatial and channel information.  The spatial information of an image tells where to focus, while the channel information helps to characterize what we are interested in. Inspired by~\cite{FC4, quasi}, our network employs the Convolutional Block Attention Module (CBAM)~\cite{cbam}, which well incorporates these two types of information. 

For the color constancy problem, recall from Eq.~\eqref{eq:MI} that if many pixels in an image are of nearly identical RGB values, they will appear as an impulse when translating to the UV space. The impulse, however, may interfere with the illuminant estimation due to the chromatism caused. Nevertheless, CBAM alleviates the negative effect by assigning lower weights to the features of these regions. The same operation can also be applied to those untrustworthy features mentioned in Section~\ref{sec:edge}. In our network, we put CBAM behind the last DenseBlock, because the deeper the network layer, the more effective the output feature information will be. This is somewhat similar to~\cite{FC4}, where a confidence layer is added before the fully connected layer.
A visualization of how these features work with attention will be given in the next section. 

    
\section{Performance Analysis}
\subsection{Datasets}
In this section, we evaluate the performance of our method on two standard color constancy datasets: i) the Color Checker dataset reprocessed by Shi and Funt~\cite{Shi2000re, Shi} and ii) the NUS $8$-camera dataset from Cheng {\it et al.}~\cite{Cheng}. The former contains $568$ images taken from two cameras, while the latter has $1736$ images (of larger size) obtained by $8$ different cameras, each of which takes about $220$ ones. For images from both datasets, the Macbeth Color Inspector (MCC) is utilized to capture the ground-truth, and the datasets provide the corners of the MCC. By setting $I_c = (0, 0, 0)$ to mask the MCC, we train and test the rest of areas in the image, which are not otherwise specially processed.


\begin{table*}[h]
            \centering
                        \caption{Performance of the ADCC variants on the reprocessed Color Checker dataset~\cite{Shi2000re}, where the best results are highlighted in \textcolor{red}{red}.}
            \begin{tabular}{cc|ccccccc}
                \hline 
& \textbf{DenseNet-$121$}&\textbf{Mean}&\textbf{Med.}&\textbf{Tri.}&\makecell{\textbf{Best} \\ \textbf{25\%}}& \makecell{\textbf{Worst} \\ \textbf{25\%}} &\makecell{\textbf{95\%} \\
\textbf{Quant.}}\\
\hline 
1 & $p=16$ & 1.99 & 1.39 & 1.52 & 0.45 & 4.54   & 5.64 \\
2 &  CBAM, $p=16$ & 1.92  & 1.23  & 1.36  & 0.37  & 4.66   & 5.97  \\
3 &  CBAM, $p=16$, Edges in FFCC\cite{FFCC} & 1.79  & 1.14  & 1.28  & 0.34  & 4.32   & 5.27 \\ 
4 &  CBAM, $p=0$, $\sigma=1/\sqrt{2}$ & 1.83  & 1.16  & 1.27  & 0.31  & 4.51 & 6.15 \\
5 &   CBAM, $p=8$, $\sigma=1/\sqrt{2}$& 1.75  & 1.12  & 1.25  & 0.34  & 4.20  & \textcolor{red}{5.03} \\
6 &   CBAM, $p=16$, $\sigma=1/\sqrt{2}$& \textcolor{red}{1.74}  & 1.13  & \textcolor{red}{1.23}  & \textcolor{red}{0.27}  & \textcolor{red}{4.17}& 5.26 & \\
 8 &  CBAM, $p=64$, $\sigma=1/\sqrt{2}$& 1.75  & 1.13  & 1.25  & 0.34  & 4.24   & 5.30  \\
9 &  CBAM, $p=16$, $\sigma=0$& 1.79  & 1.16  & 1.26  & 0.34  & 4.35 & 5.69  \\
10 &   CBAM, $p=16$, $\sigma=1$ & 1.75  & \textcolor{red}{1.08} &  1.24  & 0.33  & 4.33   & 5.14 \\
11 &  CBAM, $p=16$, $\sigma=\sqrt 2$ & 1.85  & 1.24  & 1.32  & 0.32  & 4.49  & 5.79 \\
\hline
            \end{tabular} 
	\label{table1}
\end{table*}

\begin{table*}[h]
            \centering
                        \caption{Performance comparison with previous methods on the reprocessed Color Checker and NUS $8$-camera datasets~\cite{Cheng, Shi2000re}. For both datasets, the mean, median, tri-mean, best $25\%$, worst $25\%$ angular errors are used as performance metrics. In addition, the $95\%$ quantile is considered for the reprocessed Color Checker dataset~\cite{Shi2000re} only. We highlight the best result in \textcolor{red}{red} and the runner up in \textcolor{blue}{blue}.}
            \label{table2}
            \begin{adjustbox}{width=1\textwidth}
            \begin{tabular}{cc|cccccc|ccccc}
                \hline 
 && \multicolumn{6}{c|}{\bf Reprocessed Color Checker} & \multicolumn{5}{c}{\bf NUS $8$-camera} \\
\multicolumn{2}{c|}{\textbf{Method}}&\textbf{Mean}&\textbf{Med.}&\textbf{Tri.}&\makecell{\textbf{Best} \\ \textbf{25\%}}& \makecell{\textbf{Worst} \\ \textbf{25\%}} &\makecell{\textbf{95\%} \\
\textbf{Quant.}} &\textbf{Mean}&\textbf{Med.}&\textbf{Tri.}&\makecell{\textbf{Best} \\ \textbf{25\%}}& \makecell{\textbf{Worst} \\ \textbf{25\%}} \\
\hline 
\multirow{4}{*}{\makecell{\textbf{Learning-} \\ \textbf{free}}} & White-Patch~\cite{wp} & 7.55 & 5.68 & 6.35 & 1.45 & 16.12 & - &  10.62 & 10.58 & 10.49 & 1.86 & 19.45 \\
& General Gray-World~\cite{ggw}& 4.66 & 3.48 & 3.81 & 1.00 & 10.09 & - & 3.21 & 2.38 & 2.53 & 0.71 & 7.10 \\
& Cheng {\it et al.} 2014~\cite{Cheng}& 3.52 & 2.14 & 2.47 & 0.50 & 8.74 & -  & 2.92 & 2.04 & 2.24 & 0.62 & 6.61\\
& GI~\cite{findinggray}& 3.07 & 1.87 & 2.16 & 0.43 & 7.62  & - & 2.91 & 1.97 & 2.13 & 0.56 &- \\

\hline 
\multirow{14}{*}{\makecell{\textbf{Learning-} \\ \textbf{based}}}
& Intersection-based Gamut~\cite{ebg}& 4.20 & 2.39 & 2.93 & 0.51 & 10.70 & -  & 7.20 & 5.96 & 6.28 & 2.20 & 13.61 \\
& Edge-based Gamut~\cite{ebg} & 6.52 & 5.04 & 5.43 & 1.90 & 13.58 & - & 8.43 & 7.05 & 7.37 & 2.41 & 16.08  \\
& Pixel-based Gamut~\cite{ebg}& 4.20 & 2.33 & 2.91 & 0.50 & 10.72 & 14.1  & 7.70 & 6.71 & 6.90 & 2.51 & 14.05  \\
& Bayesian~\cite{Shi} & 4.82 & 3.46 & 3.88 & 1.26 & 10.49 & - & 3.67 & 2.73 & 2.91 & 0.82 & 8.21  \\
& Natural Image Statistics~\cite{nis}& 4.19 & 3.13 & 3.45 & 1.00 & 9.22 & 11.7  & 3.71 & 2.60 & 2.84 & 0.79 & 8.47 \\
& Bright Pixels~\cite{bp}& 3.98 & 2.61 & - & - & - & - & 3.17 & 2.41 & 2.55 & 0.69 & 7.02 \\
& Spatio-spectral (GenPrior)~\cite{ss}& 3.59 & 2.96 & 3.10 & 0.95 & 7.61 & -  & 2.96 & 2.33 & 2.47 & 0.80 & 6.18  \\
& Corrected-Moment~\cite{cm}& 2.86 & 2.04 & 2.22 & 0.70 & 6.34 & -  & 2.95 & 2.05 & 2.16 & 0.59 & 6.89  \\
& Regression Tree~\cite{rt}& 2.42 & 1.65 & 1.75 & 0.38 & 5.87 & -  & 2.36 & 1.59 & 1.74 & 0.49 & 5.54\\
& CNN~\cite{CNN}& 2.63 & 1.98  & - & - & - & - & - & - & - & - & -  \\
\cline{3-13}
& CCC (dist+ext)~\cite{CCC}& 1.95 & 1.22 & 1.38 & 0.35 & 4.76 &    5.85& 2.38 & 1.48 & 1.69 & \textcolor{blue}{0.45} & 5.85  \\
& DS-Net (HypNet+SelNet)~\cite{DSNet}& 1.90 &   1.12 & 1.33 &   0.31 & 4.84 & 5.99 & 2.24 &   1.46 &   1.68 &   0.48 & 6.08 \\
& FC4(AlexNet)~\cite{FC4} & \textcolor{blue}{1.77} & \textcolor{blue}{1.11} &   1.29 & 0.34 & \textcolor{blue}{4.29} & \textcolor{blue}{5.44} &   2.12 & 1.53 & \textcolor{blue}{1.67} &   0.48 & \textcolor{blue}{4.78}  \\
& FFCC~\cite{FFCC} &   1.78 & \textcolor{red}{0.96} & \textcolor{red}{1.14} &\textcolor{blue}{0.29} &   4.62 & -   & \textcolor{blue}{1.99} & \textcolor{red}{1.31} & \textcolor{red}{1.43} & \textcolor{red}{0.35} & \textcolor{red}{4.75}  \\
& Quasi-Unsupervised~\cite{quasi} & 2.91 & 1.98 & - &-  & - & -  & \textcolor{red}{1.97} & \textcolor{blue}{1.41} & - & - & -  \\
& \textbf{Proposed} & \textcolor{red}{1.74}  & 1.13 & \textcolor{blue}{1.23} & \textcolor{red}{0.27} & \textcolor{red}{4.17} &  \textcolor{red}{5.26}  &  2.28 & 1.49 & 1.77 &   0.48 &  5.21 \\
 \hline
            \end{tabular} 

\end{adjustbox}

    \end{table*}
\begin{figure}[t]
	\centering
	 \includegraphics[width=\linewidth]{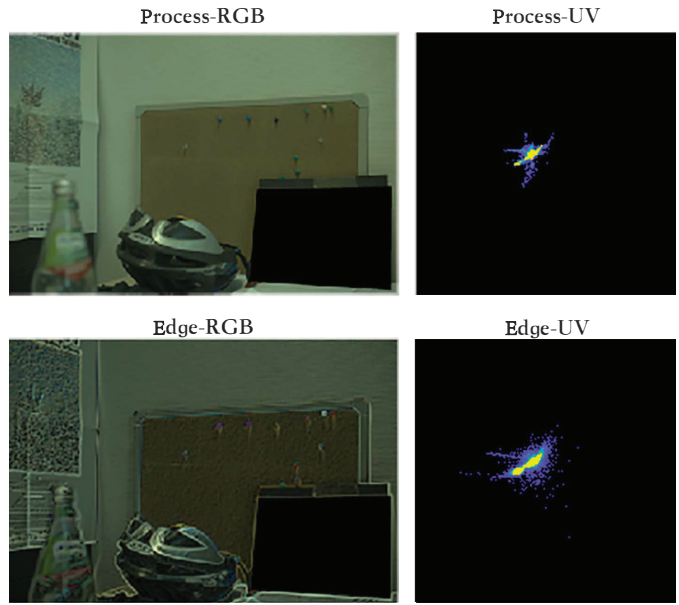}
	\caption{An example of an image after the data preprocessing.}
	\label{fig:preprocess} 
\end{figure}

\subsection{Preprocessing and Random Patches}
\paragraph{Preprocessing}
The resolution of images plays an important role in many fields of computer vision, such as image recognition~\cite{reso_recog} and segmentation~\cite{reso_seg}. For illuminant estimation, however, we mostly care about the color of images, rather than the resolution. Thus, like many learning-based color constancy methods, we downsample the images taken from high quality digital single-lens reflex  (DSLR) cameras to $256\text{px} \times 384\text{px}$ in order to accelerate our training process. Besides, following the instruction in~\cite{Shi2000re,Cheng}, we subtract the black level of cameras and abandon the pixels in images that are above the saturation level of $0.98$.  More details regarding the data processing are given as follows:
\begin{itemize}
\item Substract darkness level of each camera for each image;
\item Remove pixels that are above the saturation level;
\item Mask the color checker in the images using MCC;
\item Rotate the image if its height is larger than its width;
\item Convert images of 12-bit format to 16-bit ones;
\item Downsample images to 256\text{px} $\times$ 384\text{px};
\item Augment the edges of images with our modified Sobel filter;
\item Store the downsampled images and their augmented edges;
\item Normalize the ground-truth.
\end{itemize}
An example of an image after the data preprocessing is shown in Fig.~\ref{fig:preprocess}

\begin{figure*}[t]
\centering
\subfigure[Color Checker dataset~\cite{Shi2000re}]
{\includegraphics[width = .7\linewidth] {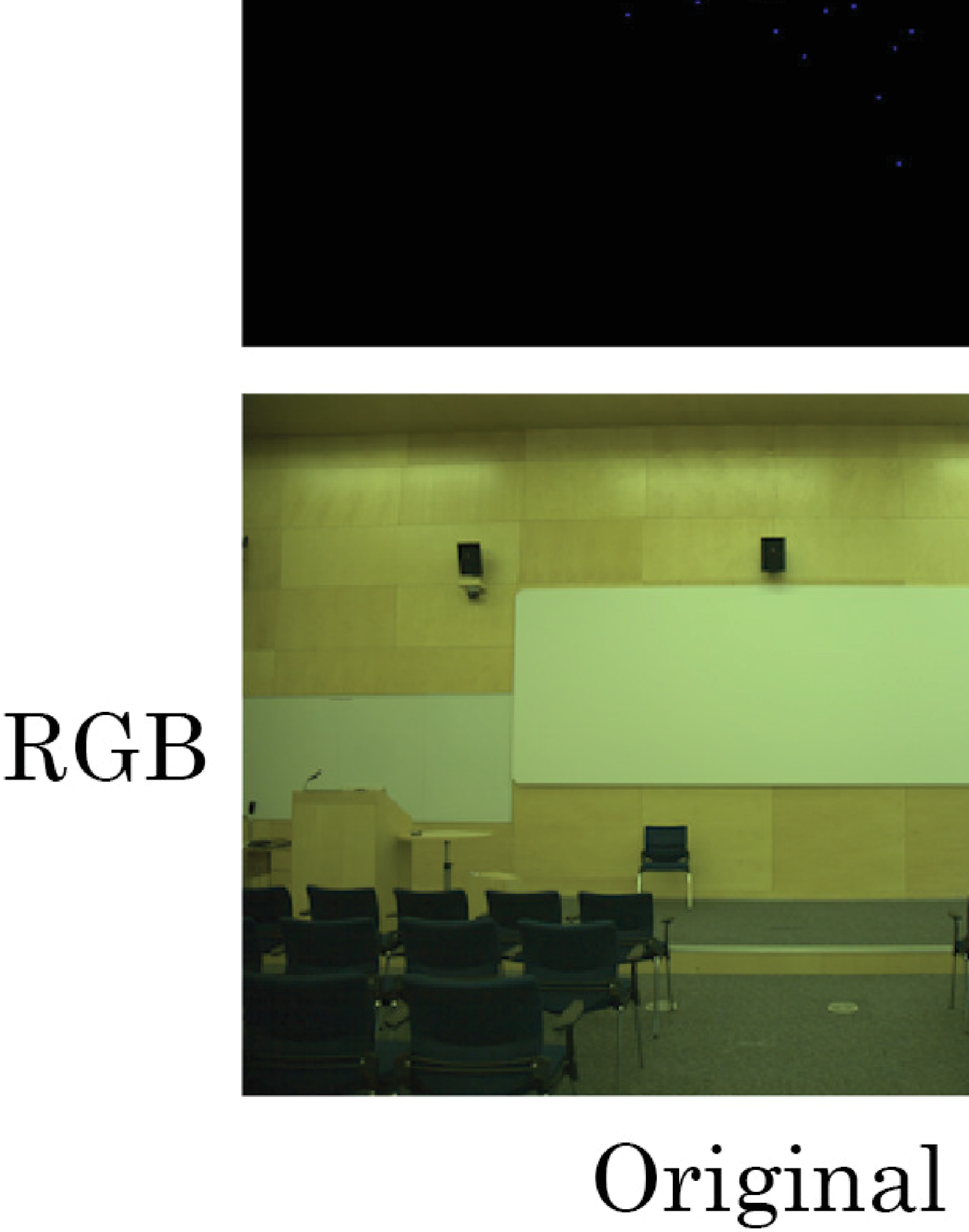}
	\label{fig:visualization} }
\subfigure[NUS $8$-camera dataset~\cite{Cheng}]
{\includegraphics[width = .7\linewidth] {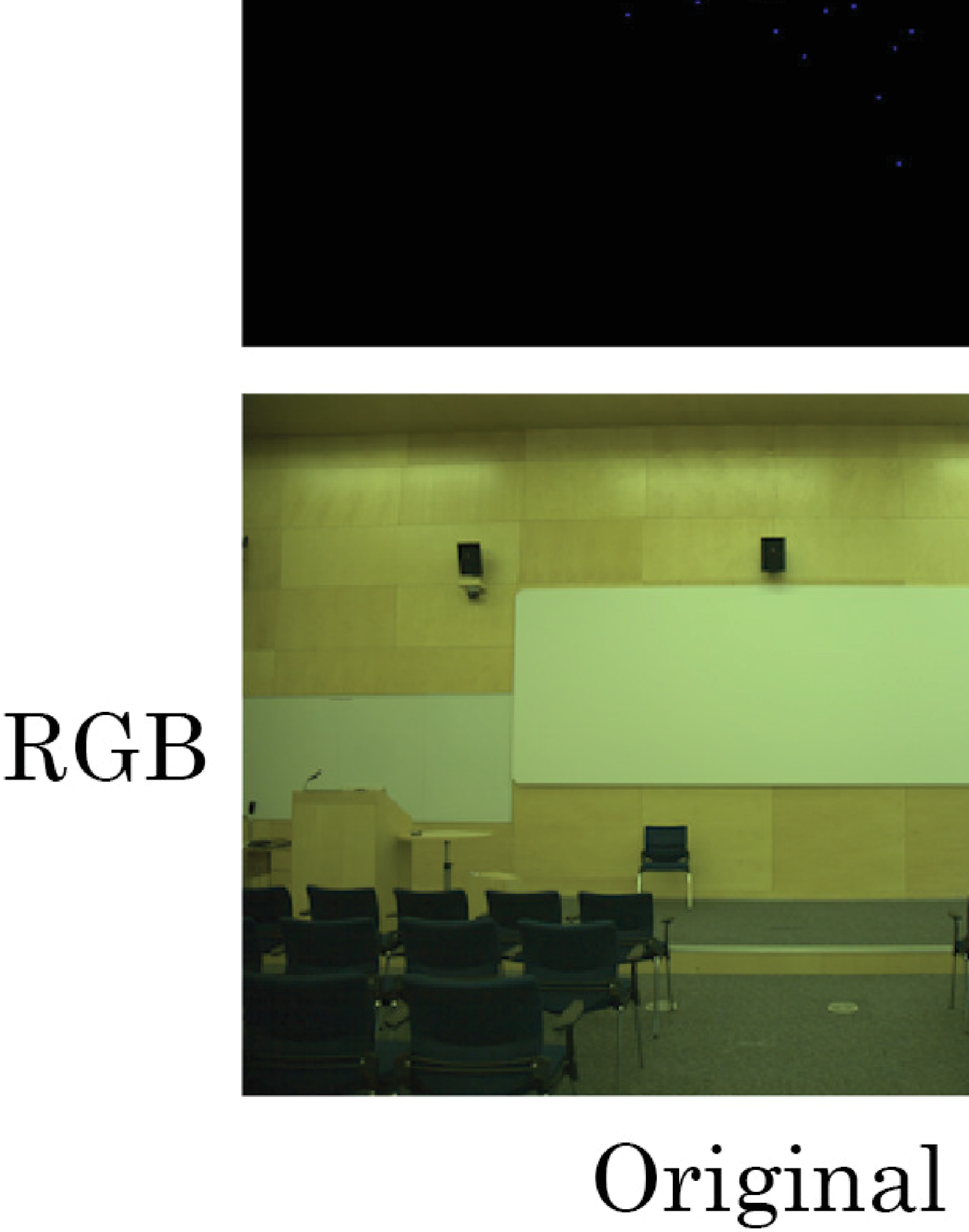} 	
   \label{fig:visualization2} }
\caption{An image from the Color Checker dataset~\cite{Shi2000re} and the model trained on the Color Checker dataset~\cite{Shi2000re} and NUS $8$-camera dataset~\cite{Cheng}, respectively.}
	\label{fig:visualization12}  
\end{figure*}

%
%
%

\begin{figure*}[t]
\centering
\subfigure[Color Checker dataset~\cite{Shi2000re}]
{\includegraphics[width = .7\linewidth] {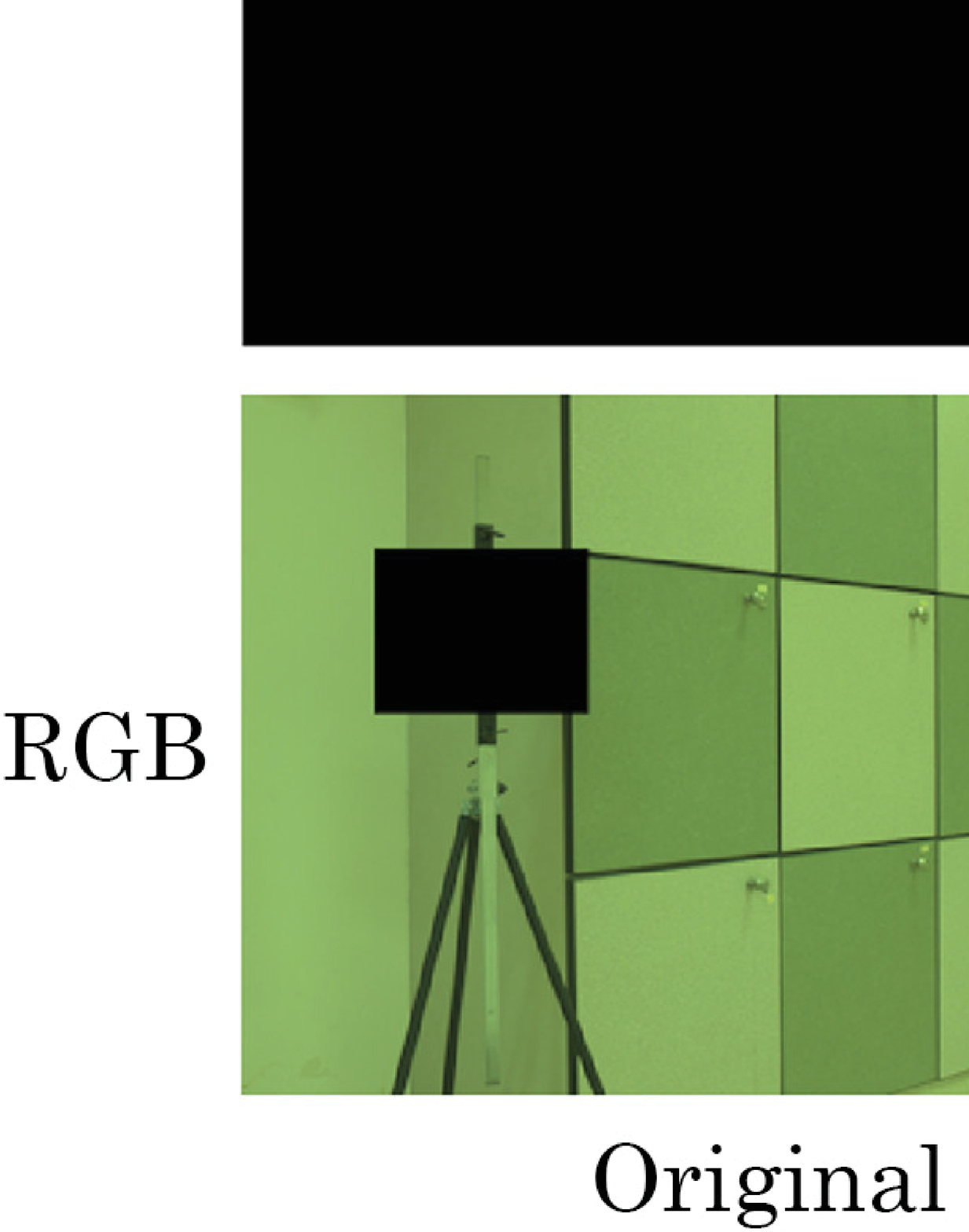}
	\label{fig:visualization3} }
\subfigure[NUS $8$-camera dataset~\cite{Cheng}]
{\includegraphics[width = .7\linewidth] {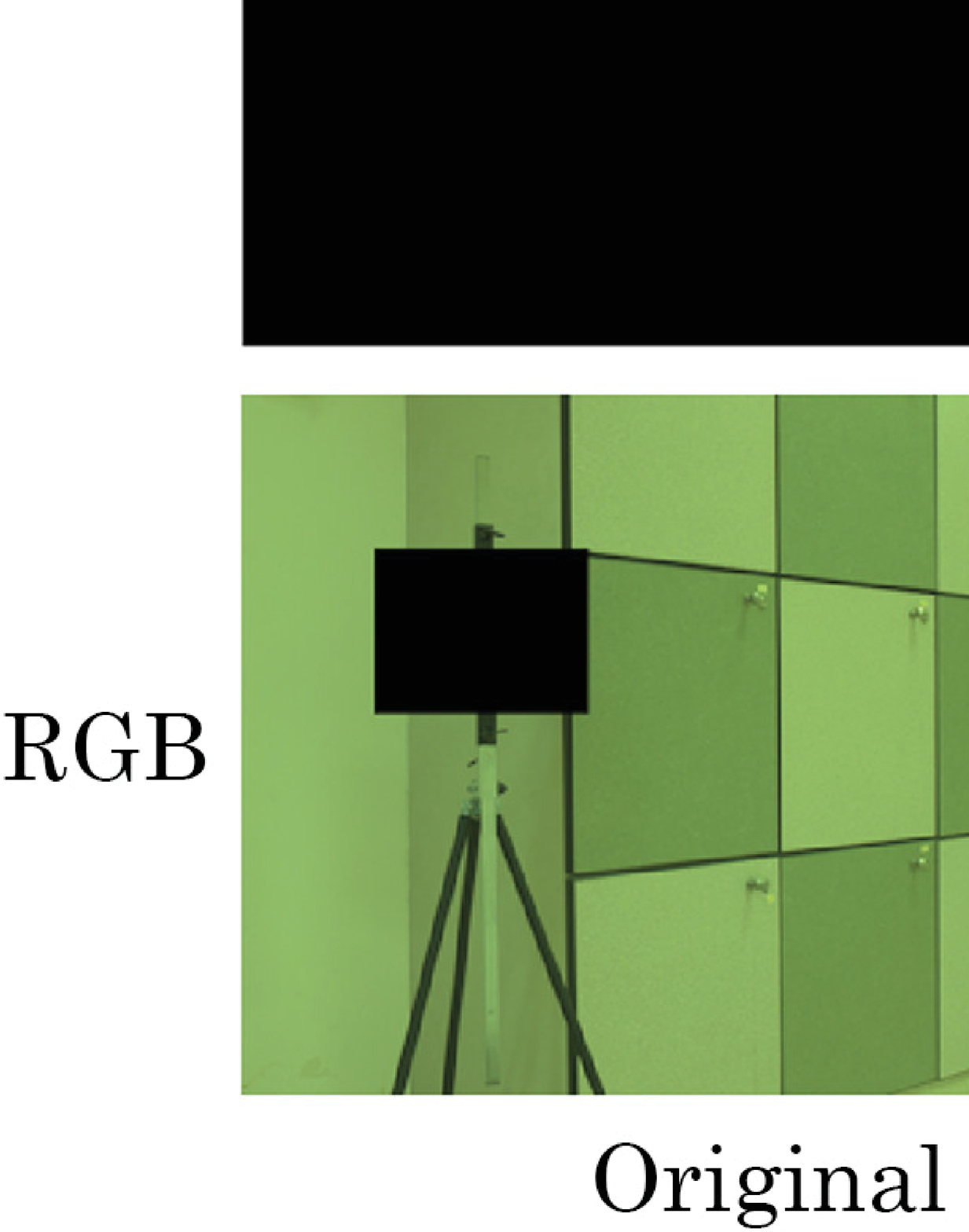} 	
   \label{fig:visualization4} }
\caption{An image from the NUS $8$-camera dataset~\cite{Cheng} and the model trained on the Color Checker dataset~\cite{Shi2000re} and NUS $8$-camera dataset~\cite{Cheng}, respectively.}
	\label{fig:visualization34}  
\end{figure*}

%
%

 \begin{figure*}[t]
	\centering
	 \includegraphics[width=.8\linewidth]{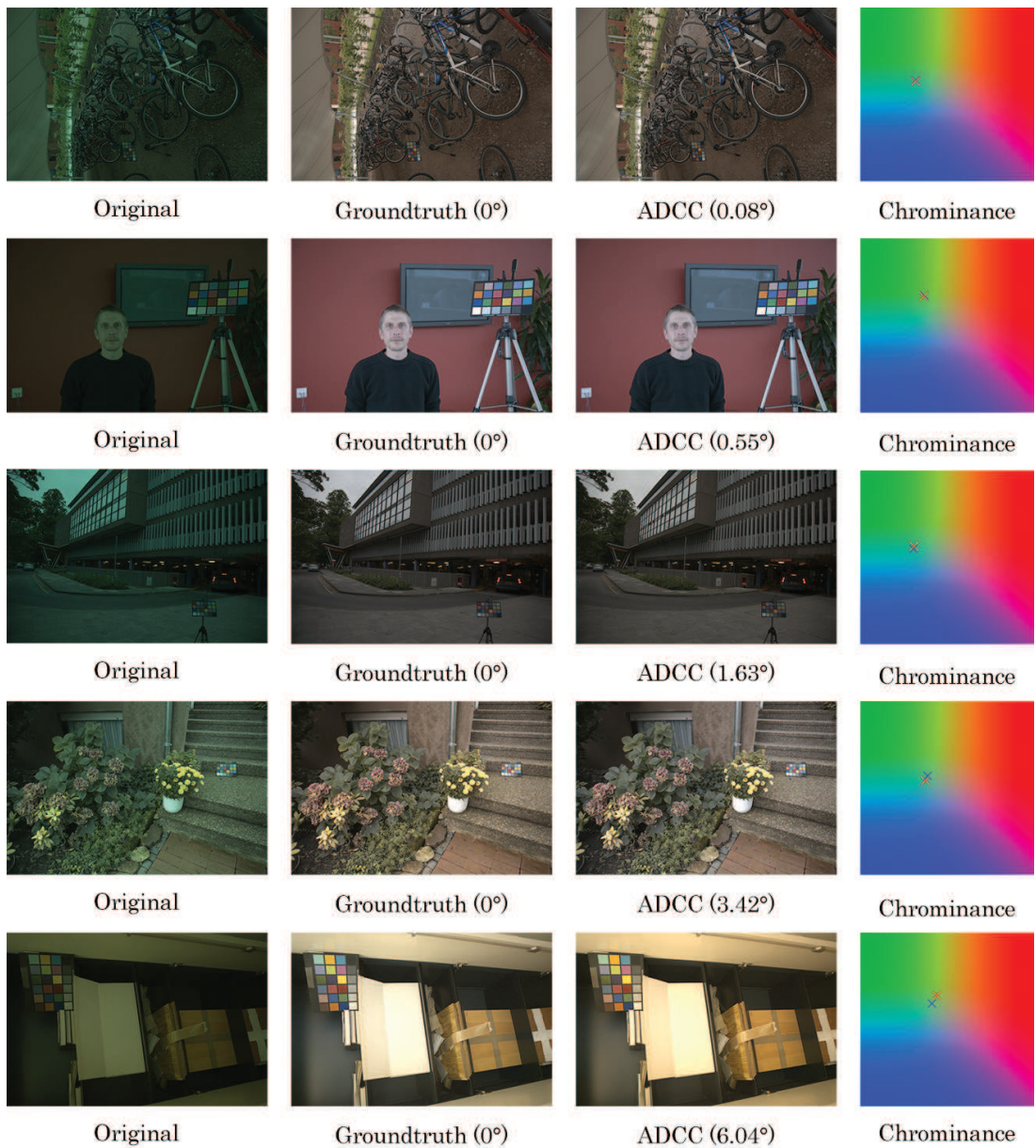}
	\caption{Examples of our results for different angular errors. From left to right: the original RAW images, images corrected with the ground-truth, images corrected with predictions by ADCC, and chrominance.}
	\label{fig:results}
\end{figure*}

\begin{figure}[t]
 \centering
  \includegraphics[width=1.0\linewidth]{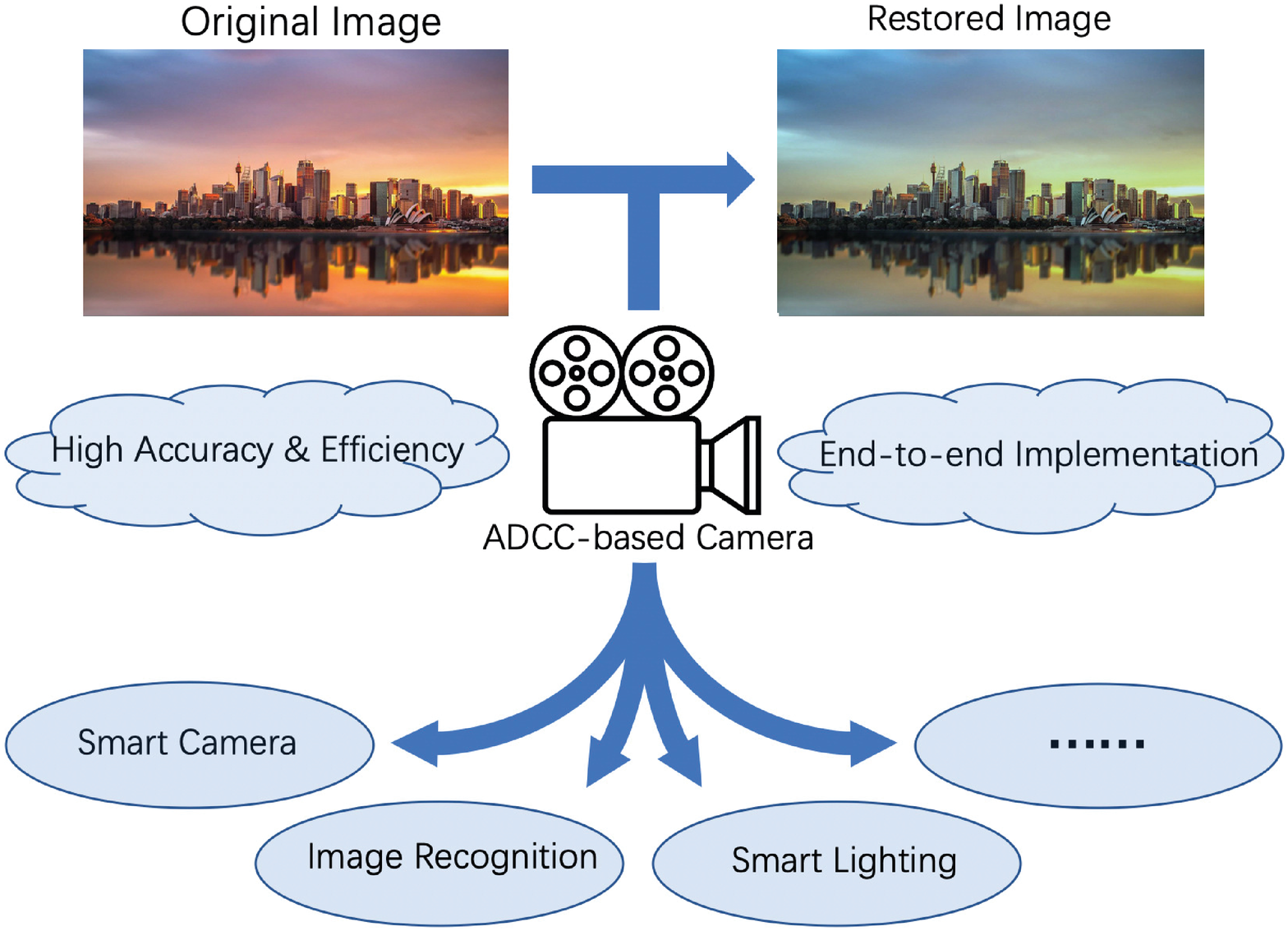}
 \caption{An illustrative example of applications where ADCC can contribute}
 \label{fig:sc}
\end{figure}

\paragraph{Patch-based improvement}
To enlarge the datasets and at the same time increase the robustness and generalization of our network, the patch-based methods are implemented. For both the training and test sets, we randomly sample multiple patches from each image as the augmented images. Specifically, we determine the height and width of the augmented images by multiplying a random number in $\left[ 0.5, 1 \right]$ to those of the original images. Also, we randomly choose the locations for the augmented images, while ensuring that they do not cross the borders of the original image.

While the original images and their augmented patches are fed into our network with their ground-truth in the training process,  they are also used to generate the local candidates of the illuminant in the test process. From these local candidates, we finally obtain the global illuminant estimation. More specifically, we compute the channel-wise median of the local predictions, followed by a normalization, to be our global prediction. According to our experiments, however, the random patches may contain some ambiguous areas, such as the yellow wall under white illuminant or the white wall under yellow illuminant~\cite{FC4}. In this case, inaccurate illuminant prediction could be made due to these misleading areas. Nevertheless, this issue can be overcome by using the median angular error, rather than the mean angular error. As shown in Fig.~\ref{fig:median}, the median angular error can better handle the situation where there are many biased local candidates, whose distribution is left-right asymmetry around the ground-truth. Of course, there is still a trade-off between the accuracy and stability of the prediction, which can be controlled by the number of patches.

Apart from the random sampling of patches, we also randomize the color of patches through a channel-wise scaling. Each channel of the patch is multiplied with a random variable in $\left[ 0.5, 1 \right]$. The same multiplier is also applied to its ground-truth. As mentioned, the multiplicative change of the RGB color is equivalent to the linear shift of its $\log$-histogram. Thus, the color randomization also forces our network to discern the shift in the $\log$-histogram and learn the translation equivariance.


\subsection{Implementation}
The proposed ADCC method is implemented by TensorFlow\cite{tens}. We train our model on the server with GTX $2080$ Ti in an end-to-end manner. {Mini-batch gradient descent is performed with a batch size of $64$ in the framework of Adam optimizer~\cite{adam}. For training epoch setting, we set epoch of $1,000$ or more for the Color Checker dataset and set epoch $2,000$ or more for the NUS 8-Camera dataset. The results may fluctuate due to randomness in the parameter initialization, patch augmentation and batch selection. To get a more stable model, we recommend to set epoch of $1,500$ for the Color Checker dataset and epoch of $3,000$ for the NUS $8$-Camera dataset.} We set the initial learning rate to $10^{-3}$ and the learning rate decay to $0.1$, where the decay does not work until the model runs to $90\%$ of the epochs, which can slightly improve the convergence speed and stability of our model. 

In addition, the dropout layer, which is often used to avoid overfitting, is not included in our model, since there are already batch normalization layers in each dense block. In our experiments, it takes about $8$ hours for training, which may be slower than some other algorithms. Nevertheless, this is reasonable because we intend to use a more complex architecture to obtain better performance on larger datasets.

\subsection{Results}
\label{sec:results}

\paragraph{Ablation Study and Internal Comparison}
To evaluate the effectiveness of edge channel and CBAM structure in our algorithm, we conduct experiments for three cases: i) neither of edge channel nor CBAM, ii) only CBAM, and iii) both of them are used. In addition, the case where the edge augmentation operator of FFCC~\cite{FFCC} is used as an alternative to ours is tested for comparative purpose.  The experimental results are shown in rows $1$ -- $3$ and $6$ of Table~\ref{table1}.

Rows $4$--$8$ in Table~\ref{table1} are the results for fine-tuning the hyper-parameters, where $p$ is the number of random patches that tradeoffs the speed and stability. To find out the best number $p$ with respect to a constant training time that is not too long, we carry out experiments. Specifically, for different $p \in \{0, 8, 16, 32, 64\}$, we choose the edge extractor  $\sigma = \frac{\sqrt{2}}{2}$ and set the training time uniformly to $12$ hours,  rather than to the corresponding epochs. The results in Table~\ref{table1} illustrate that the best $p$ is $16$. A larger $p$ leads to underfitting, while a smaller $p$ causes larger variance. The rows~$6$ and $9$ -- $11$ in Table~\ref{table1} are used to determine the edge extractor $\sigma$ for given $p = 16$. It can be observed that the model achieves the best results when $\sigma = \frac{\sqrt{2}}{2}.$

%

\paragraph{External Comparisons}
Out of consideration for the fairness of performance comparison, existing computational color constancy algorithms usually adopt the same experimental setting to both benchmark datasets; see, e.g.,~\cite{FC4, findinggray, quasi, DSNet} and the references therein. This is because a fine-tuning of network on one benchmark dataset could not be easily generalized to the other. In this paper, therefore, we follow the same practice to compare our results with others. The results on both datasets are provided in Table~\ref{table2}. 

It can be observed that for the reprocessed Color Checker dataset~\cite{Shi2000re}, ADCC performs the best among all algorithms. In particular, ADCC makes much improvement for the worst-case metric (i.e., the worst $25\%$), which clearly demonstrates the stability of the ADCC algorithm. The improvement can be attributed to the following reasons. Firstly, the reweighted feature map by CBAM effectively reduces the effect of noisy pixels in the image. Secondly, the random patches can help to reduce the variance of estimation, while using the median of local candidates as the global estimation can lower the bias of estimation. 
For the NUS $8$-camera dataset~\cite{Cheng}, the performance of ADCC is slightly inferior to that of the state-of-the-art methods. 
This is because ADCC employs a deep structure with a self-attention module, which both need sufficient amounts of data to train, and thus may not benefit very much from small datasets like the NUS $8$-camera one~\cite{Cheng}. To illustrate this, we leave an example of the spatial attention trained on the NUS-$8$ camera dataset~\cite{Cheng} to the supplementary material. The effect is not satisfactory compared to that on the Color Checker dataset~\cite{Shi2000re}.

\subsection{Visualization}
To see how the network recognizes the ambiguous pixels and edge gradients in the $\log$-histogram, we visualize two examples from
the Color Checker dataset~\cite{Shi2000re} and NUS $8$-camera dataset~\cite{Cheng} in Fig.~\ref{fig:visualization} and Fig.~\ref{fig:visualization3} respectively, the spatial attention trained on the Color Checker dataset~\cite{Shi2000re} to show how the features work with attention. In one figure, the first row shows the $\log$-histograms of the original image, augmented image, and rightmost attention. One can observe that the common parts of the concentrated chrominance (i.e., the yellow part in the two histograms), which corresponds to the non-edge pixels in the RGB images below, are assigned with light weights according to the attention map. On the contrary, the different parts of the two histograms, which represent the edge pixels and their gradients in the RGB images, respectively, are heavily weighted. It is worth mentioning that some other pixels in the image from Fig.~\ref{fig:visualization}, e.g., the chairs, are marked as ``ambiguous pixels`` as well. This is mainly due to that the shadows of an image often play a less important role than the highlights do in illuminant estimation~\cite{bp}.

For the reprocessed Color Checker dataset, ADCC performs the best among all the algorithms, while for the NUS 8-camera dataset, the performance is slightly inferior to that of the state-of-the-art methods. This is because ADCC employs a deep structure with a self-attention module, which both need sufficient amounts of data to train, and thus may not benefit very much from small datasets like the NUS $8$-camera one. To illustrate this, we use the same examples of comparison about the spatial attention trained on the NUS-$8$ camera dataset in Fig.~\ref{fig:visualization2} and Fig.~\ref{fig:visualization4}. Obviously, the attention trained on the NUS-$8$ camera dataset shows a weaker ability in recognizing the ambiguous pixels and edge gradients in the log-histogram.

In Fig.~\ref{fig:results}, we visualize  i) the original RAW images, ii) the restored images from the ground-truth illuminant, iii) the restored images based on the ADCC prediction, and iv) the corresponding chrominance. In the  chrominance column, the red crosses signify the ground-truth, while the blue ones represent our prediction, where a Gamma correction with $\gamma = \frac{1}{2.2}$ is applied to the restored RGB images for display purpose.

\subsection{Exemplary Application}
In recent years, the ADCC algorithm has rich applications in many visual fields, such as image recognition, detection, and classification. Since ADCC essentially restores the color of  images under the standard light source, it can naturally be embedded into the camera to eliminate the chromatic aberration of images. We illustrate this with an example in Fig.~\ref{fig:sc}. After being taken by camera sensors,  a picture is processed by the embedded ADCC algorithm of the camera, yielding the achromatic picture. The color constancy process can be performed in real time if needed. The high accuracy and efficiency and the end-to-end implementation of ADCC ensure the feasibility of the proposed scheme. Also, ADCC can  contribute to many other related scenarios of smart city, such as object recognition, smart lighting, and video surveillance for smart committees~\cite{liu2020semi,xia2018secure}. 

\section{Conclusion Remarks and Future Work}
In this paper, we have developed a learning-based color constancy algorithm called ADCC, which has two important features. First of all, an effective edge augmentation is used to well capture the spatial and gradient information of edges in an image. Secondly, CBAM is employed to reduce the ambiguity in the edge augmentation and feature extraction. We have demonstrated from experiments that the proposed ADCC algorithm achieves the state-of-the-art illuminant estimation performance on the reprocessed Color Checker dataset~\cite{Shi2000re}. 

For the future work, our network may not exhibit very promising performance on small datasets because of the deep structure. Nevertheless, pruning the complex structure could significantly improve the generalization of our algorithm and meanwhile shorten the training time. Furthermore, enhancing the physical portability of our algorithm to devices of limited computational resources can also be of great importance to many applications of smart cities. In addition, generalizing our algorithm to broader scenarios such as illuminant estimation under multiple light sources will also be an important direction.

\end{document}